\documentclass[10pt,letterpaper]{article}
\usepackage{wrapfig}
\usepackage{changepage}
\usepackage{graphicx}
\usepackage{tikz}
\begin{document}
\date{}
\title{\fontsize{12}{16}\selectfont{\textrm{\textbf{COVID-19 South African Vaccine Hesitancy\\ Models Show Boost in Performance Upon\\ Fine-Tuning on M-pox Tweets
}}}}
\maketitle
\author{\fontsize{8}{10}\selectfont{\textrm{\textbf{Nicholas Perikli$^{1}$}, \thanks{ \textit{School of Physics, Institute for Collider Particle Physics, University of the Witwatersrand, Johannesburg, Wits 2050, South Africa;}}}}\and
\fontsize{8}{10}\selectfont{\textrm{ \textbf{Srimoy Bhattacharya}
\footnotemark[1];}}
\and 
\fontsize{8}{10}\selectfont{\textrm{ \textbf{Blessing Ogbuokiri$^{2}$},
\thanks{ \textit{School of Applied mathematics, Africa-Canada Artificial Intelligence and Data Innovation Consortium (ACADIC), York University, Toronto, 43964, Canada;}}}}
\and 
\fontsize{8}{10}\selectfont{\textrm{ \textbf{Zahra Movahedi Nia}
 \footnotemark[2];}}
\and
\fontsize{8}{10}\selectfont{\textrm{ \textbf{Benjamin Lieberman} 
\footnotemark[1];}}
\and
\fontsize{8}{10}\selectfont{\textrm{ \textbf{Nidhi Tripathi} 
\footnotemark[1];}
\and
\fontsize{8}{10}\selectfont{\textrm{ \textbf{Salah-Eddine Dahbi} 
\footnotemark[1];}}} 
\and 
\fontsize{8}{10}\selectfont{\textrm{ \textbf{Finn Stevenson$^{3}$}, 
\thanks{ \textit{School of Physics, iThemba LABS, National Research Foundation (NRF), PO Box 722, Somerset West 7129, South Africa;}
}}} 
\and
\fontsize{8}{10}\selectfont{\textrm{{\textbf{Nicola Braggazi}
\footnotemark[2];}}}
\and
\fontsize{8}{10}\selectfont{\textrm{ \textbf{Jude Kong} 
\footnotemark[2];}}
\and
\fontsize{8}{10}\selectfont{\textrm{ \textbf{Bruce Mellado}
\footnotemark[1]}}.
}
\maketitle
\begin{abstract}
\fontsize{8}{10}\selectfont{{\textrm{
Very large numbers of M-pox cases have, since the start of May 2022, been reported in non-endemic countries leading many to fear that the M-pox Outbreak would rapidly transition into another pandemic, while the COVID-19 pandemic ravages on. Given the similarities of M-pox with COVID-19, we chose to test the performance of COVID-19 models trained on South African twitter data on a hand-labelled M-pox dataset before and after fine-tuning. More than 20k M-pox related tweets from South Africa were hand-labelled as being either positive, negative or neutral. After fine-tuning these COVID-19 models on the M-pox dataset, the F1-scores increased by more than 8$\%$ falling just short of 70$\%$, but still outperforming state-of-the-art models and well-known classification algorithms. An LDA-based topic modelling procedure was used to compare the miss-classified M-pox tweets of the original COVID-19 RoBERTa model with it’s fine-tuned version, and from this analysis we were able to draw conclusions on how to build more sophisticated models.
}}\\\\\fontsize{8}{10}\selectfont{{\textrm{\textit{{Keywords: COVID-19; Machine Learning; Natural Language Processing.}}}}}}
\end{abstract}

\section{\textbf{\fontsize{10}{13}\selectfont{{\textrm{{Introduction}}}}}}
\fontsize{10}{13}\selectfont{{\textrm{As the COVID-19 pandemic continues on, another viral disease has captured the world's attention: M-pox. Since early May 2022, cases of M-pox have been reported from countries where the disease is not endemic, and continue to be reported in several endemic countries $^{1}$. Most confirmed cases with travel history reported travel to countries in Europe and North America, rather than West or Central Africa where the M-pox virus is endemic. This is the first time that so many M-pox cases and clusters have been reported outside of endemic countries.}}}
\fontsize{10}{13}\selectfont{{\textrm{This understandably sparks the question: Do global M-pox outbreaks mark the beginning of another full-blown pandemic? M-pox and COVID-19 are both zoonotic diseases, meaning they are transmitted from animals to humans, but structurally speaking SARS-CoV-2 (the cause of COVID-19) and M-pox virus (MPV) are very different i.e., SARS-CoV-2 is an enveloped single-stranded (ribonucleic acid) RNA virus, while MPV is a double-stranded (deoxyribonucleic acid) DNA virus. Both viruses have surface proteins that facilitate their entry into host cells; however DNA viruses, like MPV, do not mutate as freely as RNA viruses, such as SARS-CoV-2.$^{2}$}}}

\fontsize{10}{13}\selectfont{{\textrm{SARS-CoV-2 is a respiratory virus, while MPV is not and primarily spreads through direct and prolonged contact with M-pox rash, scabs or body fluids from an infected person.$^{3, 4}$ Furthermore, although MPV is transmissible through sexual contact, it is not considered as a sexually transmitted disease or STD. What is clear is that, because MPV spreads primarily through close, prolonged contact, M-pox is far less transmissible than COVID-19.$^{4}$}}}

\fontsize{10}{13}\selectfont{{\textrm{COVID-19 symptoms appear anywhere from 2 to 14 days after exposure to SARS-CoV-2. Symptoms specific to COVID-19 include the general symptoms of a respiratory infection, but include the loss of taste and smell. People usually feel better after a few days to few weeks, though some people have prolonged symptoms that continue for 3+ months (i.e., long COVID) $^{3}$ However, for M-pox, it can take up to 3 weeks after exposure to MPV for symptoms to develop. Though it varies on a case-by-case basis, symptoms may mirror those of COVID-19 during the early stages of infection (e.g., fever, headache, chills).}}}

\fontsize{10}{13}\selectfont{{\textrm{Clinically speaking, M-pox differs from COVID-19 in that it is characterized by the development of painful and itchy rashes all over the body. Most people recover from M-pox after 2-4 weeks. It can be severe, even fatal, but the mortality rate is nowhere near that of COVID-19.$^{5}$ Since the beginning of 2020, COVID-19 has caused over 6,400,000 deaths across the world, though rate of deaths has declined, in part due to the availability of vaccines and treatments.$^{6}$ Risk for severe COVID-19 depends on several factors, including the variant involved, vaccination status, age and health status.$^{6}$}}}

\fontsize{10}{13}\selectfont{\textrm{According to WHO,  since the beginning of January 2022 up until mid-August there were only 12 deaths from M-pox i.e., 691 cases/day and 38 deaths/million cases, compared to the 967,282 deaths from COVID-19 i.e., 6,7141,902 cases/day and 326 deaths/million cases. Disease severity is tied, in part, to the strain of MPV causing infection. Like COVID-19, M-pox severity also depends on the same risk factors.$^{7}$}}

\fontsize{10}{13}\selectfont{{\textrm{Due to rapid antigen tests, people can test themselves for COVID-19 at home, and polymerase change reaction (PCR) testing in the lab., is also available. Both are quick and fairly accurate testing methods. However, currently, there are fewer options for diagnosing M-pox, and although PCR tests are available, there are currently no options for testing at home or at point-of-care facilities.$^{3,8}$}}}

\fontsize{10}{13}\selectfont{{\textrm{There were no vaccines for COVID-19 at the beginning of the pandemic because SARS-CoV-2 was a novel virus when it was discovered in late 2019. However, several vaccines have been developed and many have been approved for use world-wide such as AstraZeneca, Pfizer and Johnston$\&$Johnston which protects against severe disease and hospitalization, some of which are safe for persons as young as 6 months old and older. Unlike COVID-19, there is a long history of safe and effective vaccines offering protection against M-pox, which were developed for smallpox.$^{9, 10}$}}}

\fontsize{10}{13}\selectfont{{\textrm{A live-attenuated vaccine, trademarked JYNNEOS, is currently being used for widespread vaccination efforts. JYNNEOS was developed to prevent smallpox and is also protective against M-pox in adults 18 years and older. However, as of the 9$^{th}$ August, 2022, people younger than 18 years old, and at high risk for M-pox infection, may also receive the vaccine under an Emergency Use Authorization.$^{9}$}}}

\fontsize{10}{13}\selectfont{{\textrm{While JYNNEOS is the preferred vaccine for M-pox, according to the Centers for Disease Control and Prevention (CDC), there is a second smallpox vaccine, ACAM2000, that may be used as an alternative. ACAM2000 can be considered for people 1 year of age and older. However, because ACAM2000 has the potential for more adverse side effects, particularly people with weakened immune systems.$^{5, 9}$}}}

\fontsize{10}{13}\selectfont{{\textrm{M-pox vaccination efforts are currently focused on people who have been exposed to M-pox or who are more likely to get M-pox, such as 
sexually active homo-sexual men. Right now, there are no specific treatments for M-pox. However, tecovirimat, a drug that treats smallpox, may be considered for people with, or at risk for, severe disease. For both COVID-19 and M-pox, isolating infected individuals and maintaining proper hygiene (i.e., hand-washing) and disinfection practices are important for preventing and slowing the spread of infection.$^{3, 4}$}}}

\fontsize{10}{13}\selectfont{{\textrm{To answer the question, as to whether or not we are heading towards a M-pox pandemic, there are several important differences between M-pox outbreaks and the COVID-19 pandemic that we need to take into account. For one, SARS-CoV-2 was a novel virus when it emerged in late 2019, meaning it had never been seen before. As a result, the world didn’t have vaccines or immunity to the virus, which allowed it to spread like wildfire. The rise of new SARS-CoV-2 variants, coupled with the virus’s ability to transmit efficiently from person to person through the air, only fueled (and continues to fuel) this fire.$^{3}$}}}

\fontsize{10}{13}\selectfont{{\textrm{Secondly, M-pox is not a new disease. Scientists know more about MPV than they did about SARS-CoV-2 at the beginning of the COVID-19 pandemic. Importantly, given that MPV spreads primarily through close contact, it is less efficient at spreading between humans. Vaccines are also already available and, despite there having been supply chain challenges, are being administered to at-risk communities. Still, MPV is spreading in ways not previously seen (i.e., through sexual networks). The virus has also acquired mutations with unclear function and significance—if MPV continues to circulate, it could develop mutations that help it better infect humans. As such, the world must remain vigilant and put the lessons learned from the COVID-19 pandemic to good use.$^{1, 4}$}}}

\fontsize{10}{13}\selectfont{{\textrm{Given our previous success in significantly detecting and identifying COVID-19 vaccine hesitancy of South Africans from their tweets via the use of NLP models that we constructed and/or fine-tuned, and given that our fine-tuned BERT and RoBERTa models achieved the highest overall accuracies and F1-scores which were 60$\%$ and 61$\%$ respectively $^{11}$, noting that these transformer models are applicable to downstream tasks, as well as noting the high degrees of similarities and equally vast amount of differences in the structure, formation, mutation, testing, infectivity, lethality, treatment and prevention etc of the two viruses - MPV and coronavirus - we wanted to investigate how the two transformer models that we fine-tuned on COVID-19 data would behave when tested on hand-labelled sentiments pertaining to M-pox before and after fine-tuning on the M-pox dataset.}}}

\fontsize{10}{13}\selectfont{{\textrm{This study is not motivated to probe and investigate the dynamics of or correlations between vaccination hesitancy and public outcry of the M-pox outbreak in South Africa for providing immediate instruction on how to deal with these public health issues, but rather to pave the way towards the formation of more sophisticated models i.e., which could be extended for inclusion of multiple use-cases highly applicable and impactful for use in the health-care sector such as general disease detection, pandemic modelling, vaccine hesitancy identification, tracking of public opinion and mental health status during a major health threat.
}}}

\section{\fontsize{10}{13}\selectfont{{\textbf\textrm{{Materials and Methods}}}}}
\fontsize{10}{13}\selectfont{{\textrm{A total of 30000 tweets were collected using the Twitter Research License. The extraction focused on hashtags relating to M-pox over a time period spanning from the 5$^{th}$ September 2022 to the 1$^{st}$ May 2022 - which is the period before and after the peak in global infections. Duplicate tweets were dropped, leaving behind 20604 unique tweets. Unlike the previous COVID-19 study$^{11}$ no pre-processing was performed prior to training, as BERT and RoBERTa have their own built-in Tokenization schemes unique to the particular model. The COVID-19 BERT and COVID-19 RoBERTa models were tested on the M-pox dataset, then trained and tested on the M-pox dataset. The results were compared in order to test the hypothesis of boosted performance upon fine-tuning.}}}

\fontsize{10}{13}\selectfont{{\textrm{Other sets of pre-trained models were tested on the M-pox dataset with the hope of showing that hand-labelling is superior to automated labelling and that our models are unique in comparison. Thereafter, we performed LDAs on the mis-classified M-pox tweets of the COVID-19 RoBERTa model and the fine-tuned COVID-19 RoBERTa model in order to identify and then interpret the changes in the leading topics of these misclassifications pre- and post-training on the M-pox dataset - using the same procedure as in the previous paper.$^{11}$ The aim of this was to determine which of the topics present in the sample space of model misclassifications obtained pre-training remained or disappeared, and which topics emerged in the sample space of post-training model misclassifications.}}}
\subsection{\textbf{\textit{\fontsize{10}{13}\selectfont{{\textrm{Hand-Labelling of Tweets}}}}}}

\fontsize{10}{13}\selectfont{{\textrm{{The distribution of sentiments in our hand-labelled dataset were as follows: 22.2$\%$ positive; 35.3$\%$ neutral; 42.5$\%$ negative. The same hand-labelling rules and procedures were used as was followed in the previous study.$^{11}$ For this study, a collection of 20604 tweets were selected to be hand-labelled. A label is ascribed to the tweet based on the opinion of the author towards a particular theme or topic - in the previous study $^{11}$, the topic was vaccination and the theme was hesitancy, however in this case the topic was M-pox and the theme was the public's reaction to the outbreak of M-pox both in South Africa and abroad.}}}}

\fontsize{10}{13}\selectfont{{\textrm{As usual, each tweet was hand-labelled into one of three sentiment classes i.e., positive, negative, or neutral, but the criteria for hand-labelling involved answering a different question: ``Does the author of this comment feel threatened by the spread of M-pox and/or disappointed in the government's response in mitigating the spread and to what extent is he optimistic or pessimistic about the future in terms of the socio-economic stability as well as the sustainability of the mental and physical health of both himself and the community in which he resides?"}}}

\fontsize{10}{13}\selectfont{{\textrm{The three classes of sentiments were defined as follows i.e., a negative sentiment was defined as an overwhelming and irrational feeling of fear and impending doom accompanied by distrust or poor faith in the government's ability in controlling the M-pox Outbreak; a positive sentiment was defined as the absence of unwarranted fear accompanied by strong belief and deep trust in the government's ability in controlling the M-pox Outbreak; while a neutral sentiment was defined as the refusal to engage in discussions pertaining to the threat of M-pox as a public health hazard and the possible threat of a new pandemic, due to indifference, disinterest or an indecisive temperament towards the severity and planned mitigation of the M-pox Outbreak.}}}

\fontsize{10}{13}\selectfont{{\textrm{{{It must be stated that the notion of vaccine hesitancy is still applicable in this use case when labels are being chosen - since the compliance or defiance of the public towards the government's public health management may be reflected in their inclination towards vaccination in the event of a M-pox pandemic.}}}}}

\fontsize{10}{13}\selectfont{{\textrm{Additionally, a statement in which the author's viewpoint is unclear or unrelated to M-pox is by default labelled as neutral. Hence, we argue that the practise of hand-labelling is superior to automated classification algorithms - which frequently mislabel text that contain certain tones and contexts especially when negations, colloquial slang, emojis and sarcasm are present - refer to Table 2 under section A1 of the Appendix.}}}

\section*{\textbf{\fontsize{10}{13}\selectfont{{\textrm{{Results and Discussion}}}}}}
\subsection*{\textbf{\textit{\fontsize{10}{13}\selectfont{{\textrm{Machine Learning Models}}}}}}
\fontsize{10}{13}\selectfont{{{\textrm{In the previous study $^{11}$ that we performed, all of the selected machine learning models underwent extensive hyperparameter tuning using Bayesian optimization. The hyperparameters chosen for tuning the original BERT-base-cased and RoBERTa-base models on the COVID-19 data, in this previous study, were the learning rate, batch size and number of epochs - with the weight decay set to zero. These hyperparameters and their optimized values were kept when fine-tuning on the M-pox dataset. Refer to Table I, below. The overall and individual F1-scores were chosen as the defining measures for which the models could be assessed - which is also valid in the case of the M-pox dataset.}}}}

\begin{table}[ht]
\begin{tabular}{ccccccccc}
            \hline
		&  \multicolumn{4}{c}{\footnotesize{NLP-Town BERT}} &  \multicolumn{4}{c}{\footnotesize{Cardiff-NLP RoBERTa }}\\ 
		\hline
		\scriptsize Class & \scriptsize{Negative} & \scriptsize{Neutral} & \scriptsize{Positive} & \scriptsize{All} & \scriptsize{Negative} & \scriptsize{Neutral} & \scriptsize{Positive} & \scriptsize{All} \\
		\scriptsize Precision & \scriptsize 48  & \scriptsize 40  & \scriptsize 41  &  \scriptsize 43 & \scriptsize 46  & \scriptsize 38 & \scriptsize 37 &  \scriptsize 40  \\
		\scriptsize Recall & \scriptsize 37 & \scriptsize 56 & \scriptsize 22  &  \scriptsize 39  & \scriptsize 35  & \scriptsize  59  & \scriptsize 26  &  \scriptsize 40 \\
		\scriptsize F1-score & \scriptsize 42 & \scriptsize 47  & \scriptsize 30 & \scriptsize 40  & \scriptsize 40 & \scriptsize 46 & \scriptsize 31 &  \scriptsize 39\\
		\scriptsize Accuracy & \multicolumn{3}{c}{\scriptsize{}} & \scriptsize{39} & \multicolumn{3}{c}{\scriptsize{}} & \scriptsize{40}\\ 
        \hline
		&  \multicolumn{4}{c}{\footnotesize{COVID-19 BERT}} &  \multicolumn{4}{c}{\footnotesize{COVID-19 RoBERTa}}   \\ 
		\hline
		\scriptsize Class & \scriptsize{Negative} & \scriptsize{Neutral} & \scriptsize{Positive} & \scriptsize{All} & \scriptsize{Negative} & \scriptsize{Neutral} & \scriptsize{Positive} & \scriptsize{All} \\
		\scriptsize Precision & \scriptsize 56  & \scriptsize 43  & \scriptsize 43  &  \scriptsize 49 & \scriptsize 56  & \scriptsize 44 & \scriptsize 40  &  \scriptsize 48  \\
		\scriptsize Recall & \scriptsize 42  & \scriptsize 71  & \scriptsize 21  &  \scriptsize 47  & \scriptsize 40  & \scriptsize 72   & \scriptsize 21   &  \scriptsize 47 \\
		\scriptsize F1-score & \scriptsize 48 & \scriptsize 54  & \scriptsize 28  &  \scriptsize 45 & \scriptsize 47 & \scriptsize 54 & \scriptsize 28 & \scriptsize 45 \\
		\scriptsize Accuracy & \multicolumn{3}{c}{\scriptsize{}} & \scriptsize{47} & \multicolumn{3}{c}{\scriptsize{}} & \scriptsize{47}\\
		\hline
        		&  \multicolumn{4}{c}{\footnotesize{VADER Algorithm}} &  \multicolumn{4}{c}{\footnotesize{TextBlob Algorithm}}   \\
          \hline
		\scriptsize Class & \scriptsize{Negative} & \scriptsize{Neutral} & \scriptsize{Positive} & \scriptsize{All} & \scriptsize{Negative} & \scriptsize{Neutral} & \scriptsize{Positive} & \scriptsize{All} \\
		\scriptsize Precision & \scriptsize 45 & \scriptsize 52 & \scriptsize 50  &  \scriptsize 49  & \scriptsize 35 & \scriptsize 40  & \scriptsize 47  &  \scriptsize 41  \\
		\scriptsize Recall & \scriptsize 47  & \scriptsize 71 & \scriptsize 20  &  \scriptsize 50  & \scriptsize 37  & \scriptsize 62  & \scriptsize 16   &  \scriptsize 40 \\
		\scriptsize F1-score & \scriptsize 46 & \scriptsize 60  & \scriptsize 28  &  \scriptsize 47 & \scriptsize 36 & \scriptsize 49 & \scriptsize 23 &  \scriptsize 37 \\
		\scriptsize Accuracy & \multicolumn{3}{c}{\scriptsize{}} & \scriptsize{50} & \multicolumn{3}{c}{\scriptsize{}} &  \scriptsize{40}\\
		\hline
        		&  \multicolumn{4}{c}{\footnotesize{Fine-Tuned BERT }} &  \multicolumn{4}{c}{\footnotesize{ Fine-Tuned RoBERTa}}   \\
          \hline
		\scriptsize Class & \scriptsize{Negative} & \scriptsize{Neutral} & \scriptsize{Positive} & \scriptsize{All} & \scriptsize{Negative} & \scriptsize{Neutral} & \scriptsize{Positive} & \scriptsize{All} \\
		\scriptsize Precision & \scriptsize 69  & \scriptsize 73 & \scriptsize 60  &  \scriptsize 69  & \scriptsize 67  & \scriptsize 77 & \scriptsize 64 &  \scriptsize 70 \\
		\scriptsize Recall & \scriptsize 72  & \scriptsize 71 & \scriptsize 57  &  \scriptsize 68 & \scriptsize 77 & \scriptsize  68 & \scriptsize 57  &  \scriptsize 69\\
		\scriptsize F1-score & \scriptsize 70 & \scriptsize 72  & \scriptsize 58 &  \scriptsize 68  & \scriptsize 72 & \scriptsize 72 & \scriptsize 61 &  \scriptsize 69\\
		\scriptsize Accuracy & \multicolumn{3}{c}{\scriptsize{}} & \scriptsize{68} & \multicolumn{3}{c}{\scriptsize{}} & \scriptsize{69}\\
		\hline
        \end{tabular}
        \caption{\underline{\fontsize{8}{10}\selectfont{\textrm{{\textrm{COVID-19 vs pre-trained Model Performance on the M-pox data.}}}}}}
        \end{table}
\fontsize{10}{13}\selectfont{\textrm{{\textrm{Pre-selected BERT pre-trained models were used, whose results served as a comparison to the performance of our respective fine-tuned models. These pre-trained models were identical to those used in the previous study on COVID-19. The pre-trained NLP-Town BERT model when tested on the M-pox dataset achieved an overall precision of 43$\%$, an overall accuracy of 39$\%$ yielding an F1-score of 40$\%$, while the pre-trained Cardiff-NLP RoBERTa model achieved a similar result with an overall precision of 40$\%$, overall accuracy of 40$\%$ yielding an F1-score of 39$\%$.}}}} 

\fontsize{10}{13}\selectfont{\textrm{{\textrm{Our COVID-19 models performed significantly better than the pre-trained models on the M-pox dataset i.e., the COVID-19 BERT model  achieved an overall precision of 49$\%$, an overall accuracy of 47$\%$ yielding an F1-score of 45$\%$, while the COVID-19 RoBERTa model achieved a similar result with an overall precision of 48$\%$, overall accuracy of 47$\%$ yielding an F1-score of 45$\%$.}}}}

\fontsize{10}{13}\selectfont{\textrm{{\textrm{Pre-selected and pre-trained text classification algorithms i.e. TextBlob and VADER were also used to label the tweets in the M-pox dataset, whose labels were then tested against or checked for correlation against their corresponding hand labels. The results from the TextBlob and VADER algorithms served as a comparison for the degree of similarity or dissimilarity in the criteria used when classifying sentiments via automated means versus classifying sentiments using a manual approach. The VADER algorithm when tested on the M-pox dataset achieved an overall precision of 49$\%$, an overall accuracy of 50$\%$ yielding an F1-score of 47$\%$, while the TextBlob algorithm achieved a significantly poorer result with an overall precision of 41$\%$, overall accuracy of 40$\%$ yielding an F1-score of 37$\%$. This shows the inadequacy of automated labelling when compared to hand-labelling, as well as the deviations in agreement amongst automated labelling algorithms themselves.}}}}

\fontsize{10}{13}\selectfont{\textrm{{\textrm{After fine-tuning our COVID-19 models on the M-pox dataset, the model performance was remarkably better i.e., the fine-tuned RoBERTa  model registered an overall precision of 70$\%$, overall accuracy of 69$\%$ yielding an F1-score of 69$\%$, while, similarly, the the fine-tuned BERT model achieved an overall precision of 69$\%$, overall accuracy of 68$\%$ yielding an F1-score of 68$\%$.}}}}

\fontsize{10}{13}\selectfont{\textrm{{\textrm{Comparatively, even though the COVID-19 RoBERTa and COVID-19 BERT models performed quite poorly when tested on the M-pox dataset, they significantly outperformed the pre-trained NLP-Town BERT and Cardiff-NLP RoBERTa models, with scores ranging between 5$\%$ and 10$\%$ higher over the three overall performance measures. This is easily explained if one considers that the first set of models are largely based on the training of tweets from countries in the Global North with more general use-cases, while our models were trained on South African tweets with the specific use-case of vaccine hesitancy - in which the cultural and linguistic differences in the way people communicate, as well as differences in the themes and the topics within the text with respect to the relevant subject matter, in the two respective hemispheres clearly played a role in overall model performance.}}}}

\fontsize{10}{13}\selectfont{\textrm{{\textrm{Moreover, the fine-tuned COVID-19 BERT and roBERTa models far outperformed their original COVID-19 versions with scores ranging between 20$\%$ and 25$\%$ higher over the three overall performance measures. This too can easily be explained if one uses the same reasoning as above and also considers that there is a significant overlap or correlation between the emotional states of people when M-pox is used as the use-case and the extent of vaccine hesitancy that people displayed in the COVID-19 dataset when vaccination was used as the use-case.}}}}

\fontsize{10}{13}\selectfont{\textrm{{\textrm{In other words, these results validate our hypothesis that fine-tuning the COVID-19 models on the M-pox dataset, would improve the performance of the models in correctly classifying hand-labelled M-pox tweets with M-pox as the use-case and corresponding labels of positive, negative and neutral, which show-cases the similarities between the vaccine hesitancy and M-pox datasets in terms of linguistic style, overlapping topics as well as themes inherent in the tweets themselves. This motivates the use of topic modelling to investigate the dynamics of this overlap and the intricacies of the correlation between these two use-cases and their corresponding data-sets.}}}}
\subsubsection*{\textmd{\textit{{\textrm{Topic Modelling}}}}}
\fontsize{10}{13}\selectfont{\textrm{{\textrm{In this study, an LDA was performed on the set of tweets that were miss-classified by the fine-tuned RoBERTa model, or the COVID-19 model, post-training on the M-pox dataset - and the RoBERTa model, or the COVID-19 RoBERTa model.}}}} 

\fontsize{10}{13}\selectfont{\textrm{{\textrm{The clusters were then visualized, and the topics were identified for both of these models, respectively.  The same procedure was followed as was used in the previous study, in which the LDA was built using the Gensim module lda-model function and the number of topics was set to 5.$^{11}$}}}}

\fontsize{10}{13}\selectfont{\textrm{{\textrm{The top 30 most salient terms in each topic were extracted, leading topics visualized using the pyLDAvis tool and thereafter, the 5 topics were identified for the respective cases. The top 10 Most Frequent Terms per LDA Cluster Grouping for the two cases as well as their associated leading topics are shown in Table 3 under section A2 of the Appendix.}}}}

\fontsize{10}{13}\selectfont{\textrm{{\textrm{The close enough 5 topics for the respective cases are listed below and their respective distributions inferred from the LDA are shown in Figures 1 and 2 below}}}}:\\\\\\\\
\begin{itemize}
	\item \fontsize{10}{13}\selectfont{\textrm{Topic 1 : Vaccine Safety and Availability Concerns}}
	\item \fontsize{10}{13}\selectfont{\textrm{Topic 2 : Fear of Death from M-pox Infection}} 
	\item \fontsize{10}{13}\selectfont{\textrm{Topic 3 : Modes of M-pox Transmission}}
	\item \fontsize{10}{13}\selectfont{\textrm{Topic 4 : Conspiracy Theories about the M-pox outbreak}}
	\item \fontsize{10}{13}\selectfont{\textrm{Topic 5 : M-pox related Stigmatisation and Discrimination}}  
\end{itemize}
\begin{figure}[ht]
	\centering
\vspace*{8pt}
\includegraphics[width=1.05\textwidth,height=0.602\textwidth]{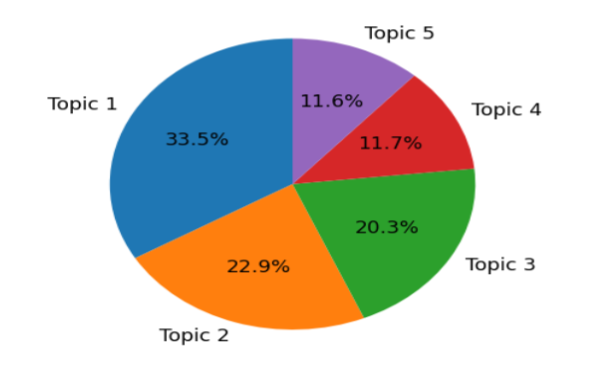}
	\caption{\label{LDATopicS}The Distribution of the general topics in the sample space of M-pox tweets miss-classified by the RoBERTa model pre-training.}
\end{figure}
\begin{itemize}
	\item \fontsize{10}{13}\selectfont{\textrm{Topic 1 : M-pox as a Sexually Transmitted Disease}}
	\item \fontsize{10}{13}\selectfont{\textrm{Topic 2 : Vaccine Safety and Availability Concerns}}
	\item \fontsize{10}{13}\selectfont{\textrm{Topic 3 : Conspiracy Theories about the M-pox Outbreak}}		
	\item \fontsize{10}{13}\selectfont{\textrm{Topic 4 : Fear of M-pox related skin lesions and scarring}}
	\item \fontsize{10}{13}\selectfont{\textrm{Topic 5 : Potential Emergence of a deadly M-pox pandemic}}  
\end{itemize}
\begin{figure}[ht]
	\centering
	\includegraphics[width=0.97\textwidth,height=0.65\textwidth]{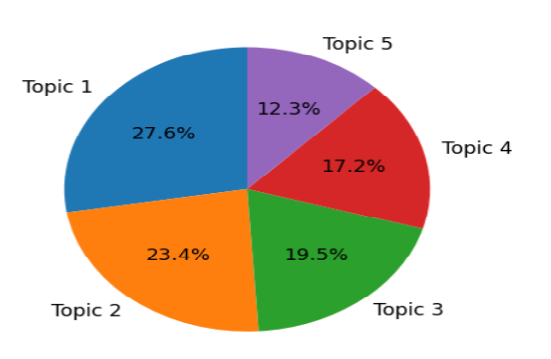}
\vspace*{8pt}
  \caption{\label{LDATopicS}{The Distribution of the general topics in the sample space of M-pox tweets miss-classified by the RoBERTa model post-training.}}
\end{figure}

\fontsize{10}{13}\selectfont{\textrm{From the pie charts in Figures 1 and 2, it is clear that only two topics survived post-training i.e., ``Vaccine Safety and Availability Concerns'' (33.5$\%$-27.6$\%$) and ``Conspiracy Theories about the M-pox Outbreak'' (11.7$\%$-19.5$\%$). The topics that disappeared in leading order were: ``Fear of Death from M-pox Infection'' (22.9$\%$), ``Possible Modes of M-pox Transmission'' (20.3$\%$) and ``M-pox-related Stigmatization and Discrimination'' (11.6$\%$), and the emergent topics in leading order were: ``M-pox as an STD" (27.6$\%$), ``Fear of M-pox related skin lesions and Scarring" (17.2$\%$) and the ``Potential Emergence of a Deadly M-pox Pandemic" (12.3$\%$)}}. 

\fontsize{10}{13}\selectfont{\textrm{{\textrm{It is apparent that although 3 out of the 5 topics in the one set of LDA results are not identical to the other, that they are in essence the same entity or grouping i.e. topics within a topic i.e.  - ``Possible Modes of M-pox Transmission'' and ``M-pox as an STD.'', which are both sub-topics of the general topic of M-pox transmission and infectivity; ``Fear of M-pox related skin lesions and scarring" and ``Fear of death by M-pox", which are both sub-topics of the general topic of M-pox infection and recovery; ``Potential Emergence of an M-pox pandemic'' and ``M-pox related Stigmatization and Discrimination'', which are both sub-topics of the general topic of public mass panic and hysteria.}}}} 

\fontsize{10}{13}\selectfont{\textrm{{\textrm{This leads to the conclusion that there are over-lapping themes in the dataset, owed to the fact that the proportions of the five topics pertaining to a general topic have shifted but not in the case of Vaccination. This phenomenon can be further investigated and the results can be used to further improve the accuracy and efficiency of the model.}}}}

\section*{\fontsize{10}{13}\selectfont{\textrm{\textbf{Conclusion}}}}
\fontsize{10}{13}\selectfont{\textrm{{\textrm{Motivated by a previous study $^{11}$, whose results were published in a paper named, ``A Natural Language Processing approach to Probe COVID-19 Vaccine Hesitancy from Tweets in South Africa", we hand-labelled twitter data using M-pox as a use-case in order to test the hypothesis that a better model performance would be obtained from the COVID-19 BERT and RoBERTa models after fine-tuning on the collection of hand-labelled M-pox tweets, than if there was no fine-tuning prior to testing.}}}}

\fontsize{10}{13}\selectfont{\textrm{{\textrm{We tested pre-trained models on the M-pox dataset. All models performed poorly in comparison to the COVID-19 models and the fine-tuned COVID-19 models, illustrating the inadequacy of automated labelling when compared to hand-labelling, the complexity in the use-case of M-pox in relation to the much simpler use-cases used in developing many pre-trained models, as well as the uniqueness of the South African dataset in terms of the use of language and communication style, as compared to other cultural groups in other parts of the world.}}}}

\fontsize{10}{13}\selectfont{\textrm{{\textrm{The additional training of our fine-tuned COVID-19 models on the M-pox dataset resulted in a significant improvement in overall performance, thus confirming our hypothesis, and hence it was concluded that since not only can our COVID-19 models perform reasonably well on detecting the presence of COVID-19 vaccine hesitancy amongst the South African republic, but also perform reasonably well on inferring the emotional states i.e., degree of optimism or despair of individuals within the South African republic with regards to the spread of M-pox within the community and the effectiveness of the control measures used in addressing the Outbreak.}}}} 

\fontsize{10}{13}\selectfont{\textrm{{\textrm{Given the set LDA results, after comparing the findings pre- and post- fine-tuning the COVID-19 models on the M-pox, it was concluded that there are over-lapping themes in the dataset, owed to the fact that the proportions of the five topics pertaining to a general topic have shifted but not in the case of Vaccination. This phenomenon can be further investigated and the results can be used to further improve the accuracy and efficiency of the model. The general topics were found to be mass hysteria, infectivity and transmission, disease and recovery, conspiracy theories, and finally the safety and availability of vaccines.}}}}

{\fontsize{10}{13}\selectfont{\textrm{It remains to be seen if this boost in performance can be extended to formulate models applicable to all types of worrisome diseases amongst all cultural groups throughout the world.}}}
\section*{\textrm{\underline{Appendix}}}
\section*{A.1}
\subsection*{\fontsize{10}{13}\selectfont{\textrm{\underline{Hand-labelling of Tweets}}}}
\vspace{-0.25cm}
\begin{table}[ht]
{\resizebox{\textwidth}{!}{
\begin{tabular}{ccccc} 
\hline
\fontsize{8}{10}\selectfont{\textrm{Cases}} & \fontsize{8}{10}\selectfont{\textrm{Tweet}} & \fontsize{8}{10}\selectfont{\textrm{Hand}} & \fontsize{8}{10}\selectfont{\textrm{VADER}} & \fontsize{8}{10}\selectfont{\textrm{TextBlob}} \\ \hline
\fontsize{8}{10}\selectfont{\textrm{}} & \fontsize{8}{10}\selectfont{\textrm{ We beat COVID, now let’s beat M-pox.}} \includegraphics[scale=0.035]{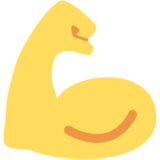} \includegraphics[scale=0.035]{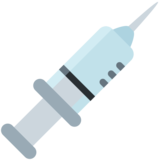} $\#$M-pox & \fontsize{8}{10}\selectfont{\textrm{+}} & \fontsize{8}{10}\selectfont{\textrm{+}} & \fontsize{8}{10}\selectfont{\textrm{+}}\\
\fontsize{8}{10}\selectfont{\textrm{ Clear-cut}} & \fontsize{8}{10}\selectfont{\textrm{M-pox is a hoax! Get real, people!}} & \fontsize{8}{10}\selectfont{\textrm{ \textbf{-}}} & \fontsize{8}{10}\selectfont{\textrm{ \textbf{-}}} & \fontsize{8}{10}\selectfont{\textrm{ 0}}\\
\fontsize{8}{10}\selectfont{\textrm{}} & \fontsize{8}{10}\selectfont{\textrm{ M-pox does not seem to be as deadly as everyone says}} & \fontsize{8}{10}\selectfont{\textrm{ 0}} & \fontsize{8}{10}\selectfont{\textrm{ 0}} & \fontsize{8}{10}\selectfont{\textrm{ 0}} \\\hline
\fontsize{8}{10}\selectfont{\textrm{}}  & \fontsize{8}{10}\selectfont{\textrm{ My family got M-pox. All had minor symptoms and are now 100$\%$.}}  & \fontsize{8}{10}\selectfont{\textrm{ +}} & \fontsize{8}{10}\selectfont{\textrm{ 0}} & \fontsize{8}{10}\selectfont{\textrm{ 0}} \\
\fontsize{8}{10}\selectfont{\textrm{ Border-line}} & \fontsize{8}{10}\selectfont{\textrm{ A M-pox Outbreak right after COVID ends? Hmm... \includegraphics[scale=0.035]{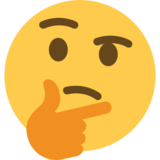}
\includegraphics[scale=0.035]{thinking-face} \includegraphics[scale=0.035]{thinking-face}}} & \fontsize{8}{10}\selectfont{\textrm{ \textbf{-}}} & \fontsize{8}{10}\selectfont{\textrm{ 0 }}& \fontsize{8}{10}\selectfont{\textrm{ 0}} \\
\fontsize{8}{10}\selectfont{\textrm{ }} & \fontsize{8}{10}\selectfont{\textrm{ Now there is M-pox. \includegraphics[scale=0.125]{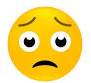} Seriously? \includegraphics[scale=0.035]{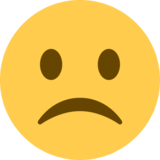} Should I be concerned? \includegraphics[scale=0.035]{thinking-face}}}  & \fontsize{8}{10}\selectfont{\textrm{ 0}} & \fontsize{8}{10}\selectfont{\textrm{ \textbf{-} }}& \fontsize{8}{10}\selectfont{\textrm{ \textbf{-} }}\\ \hline
\fontsize{8}{10}\selectfont{\textrm{}}  & \fontsize{8}{10}\selectfont{\textrm{ I'll take pre-caution, but if I get it and die, then I die. Oh well!}}  & \fontsize{8}{10}\selectfont{\textrm{ +}} & \fontsize{8}{10}\selectfont{\textrm{ \textbf{-}}} & \fontsize{8}{10}\selectfont{\textrm{ 0}} \\
\fontsize{8}{10}\selectfont{\textrm{ Difficult}} & \fontsize{8}{10}\selectfont{\textrm{ I have fully recovered from M-pox, but a friend got it and died.}} & \fontsize{8}{10}\selectfont{\textrm{ \textbf{-} }}& \fontsize{8}{10}\selectfont{\textrm{ \textbf{-} }}& \fontsize{8}{10}\selectfont{\textrm{ 0 }}\\
\fontsize{8}{10}\selectfont{\textrm{ }} & \fontsize{8}{10}\selectfont{\textrm{ M-pox can infect people of any race and any sexual orientation.}} & \fontsize{8}{10}\selectfont{\textrm{ 0}} & \fontsize{8}{10}\selectfont{\textrm{ 0}} & \fontsize{8}{10}\selectfont{\textrm{ 0}} \\ \hline
\fontsize{8}{10}\selectfont{\textrm{}} & \fontsize{8}{10}\selectfont{\textrm{ WHO is doing a great job. Let's support them!! \includegraphics[scale=0.0355]{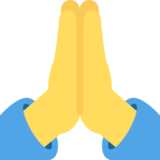} \includegraphics[scale=0.035]{folded-hands}}} & \fontsize{8}{10}\selectfont{\textrm{ +}} & \fontsize{8}{10}\selectfont{\textrm{ +}} & \fontsize{8}{10}\selectfont{\textrm{ 0}} \\ 
\fontsize{8}{10}\selectfont{\textrm{ Same Text}} & \fontsize{8}{10}\selectfont{\textrm{ WHO is doing a great job. Let's support them!!  \includegraphics[scale=0.035]{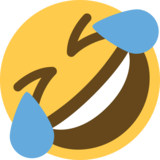} \includegraphics[scale=0.035]{rolling-on-the-floor-laughing} \includegraphics[scale=0.035]{rolling-on-the-floor-laughing} \includegraphics[scale=0.035]{rolling-on-the-floor-laughing} \includegraphics[scale=0.035]{rolling-on-the-floor-laughing}}} & \fontsize{8}{10}\selectfont{\textrm{ \textbf{-}}} & \fontsize{8}{10}\selectfont{\textrm{ +}} & \fontsize{8}{10}\selectfont{\textrm{ 0}} \\
\fontsize{8}{10}\selectfont{\textrm{}} & \fontsize{8}{10}\selectfont{\textrm{ WHO is doing a great job. Let's support them??}} & \fontsize{8}{10}\selectfont{\textrm{ 0}} & \fontsize{8}{10}\selectfont{\textrm{ +}} & \fontsize{8}{10}\selectfont{\textrm{ 0}} \\ \hline
\end{tabular}}}
\caption{\fontsize{8}{10}\selectfont{\textrm{{\underline{An illustration of the advantages of Manual over Automated text labelling.}}}}}
\end{table}
\fontsize{9}{12}\selectfont{\textrm{{\textrm{{Here we highlight the pitfalls of using text classification algorithms over hand-labelling using explicit examples. In Table 2, the four different cases of tweets one would encounter when performing sentiment analysis along with three hand-labelled examples for each case, each corresponding to one of the three sentiment classes i.e., positive (+), negative (-), neutral (0) are provided. The four categories are: clear-cut cases, borderline cases, difficult-to-label tweets and same text tweets.}}}} 

\fontsize{9}{12}\selectfont{\textrm{{\textrm{Clear-cut cases correspond to tweets whose sentiment labels are obvious and there is no debate on the validity of its classification - in other words the tweet's polarity is heavily skewed towards a single sentiment type. Borderline cases correspond to tweets that can arguably take on one of two labels i.e., either neutral or positive or alternatively neutral or negative, whereby the author's point of view is debatable. Difficult-to-label tweets are tweets that contain both positive and negative sentiments each with high polarity scores, which makes it difficult to decided on the overall text polarity. Same text tweets are a class of tweets whereby the raw text is identical but differ in the amount of punctuation and/or emojis present in the tweet, which serve to change the message behind the tweet often through the introduction of satire.}}}} 

\fontsize{9}{12}\selectfont{\textrm{{\textrm{Two different classification algorithms were selected namely, VADER and TextBlob. These classification algorithms were then given each example tweet and their predicted labels were compared to the manually-classified tweet labels. The results are presented in the table. Overall VADER correctly predicted the labels of 50$\%$ of the tweets, in which 100$\%$ of the clear-cut case examples were classified correctly, while none, or 0$\%$, of the border-line case tweets were classified correctly and only one third, 33$\%$, of the difficult-to-label or the same text tweets were correctly labelled. VADER was able to get 50$\%$ recall for each respective class. Comparatively, TextBlob correctly predicted the labels of a third, or $\approx$ 33$\%$, of all the tweets, in which two thirds, or $\approx$ 67$\%$, of the clear-cut case examples were classified correctly, while none, or 0$\%$, of the border-line case tweets were classified correctly, 33$\%$ of all of the difficult-to-label tweets were labelled correctly, but none, 0$\%$, of the same text tweets were correctly labelled. TextBlob got recalls of 25$\%$ for the positives, 75$\%$ for the neutrals but nothing, 0$\%$, for the negatives.}}}} 

\fontsize{9}{12}\selectfont{\textrm{{\textrm{This shows that both classification algorithms perform well on simple clear-cut examples, but become much less efficient in correctly classifying tweets, as the complexity of the tweets increases. Furthermore, given the recall values, it is apparent that VADER is equally good in labelling each sentiment type, while TextBlob strongly favours a neutral label. In both cases, the overall accuracies are very low in comparison to hand-labelling and it is clear that when given same text tweets, the algorithms are unable to identify sarcasm or the nuanced effect of changing punctuation marks i.e., from ! to ?, given that VADER provided a positive label for each sentiment belonging to the same text case, while TextBlob provided all neutral labels.}}}} 

\fontsize{9}{12}\selectfont{\textrm{{\textrm{Hence, the table clearly highlights the advantages of manual over automated hand-labelling. A very similar result was obtained with respect to the COVID-19 dataset.}}}}}
\section*{A.2}
\subsection*{\underline{\fontsize{10}{13}\selectfont{\textrm{Linear Discriminant Analysis}}}}
\vspace{-0.25cm}
        \begin{table}[ht]
        {\resizebox{\textwidth}{!}{
       \begin{tabular}{cccc} 
            \hline
		\fontsize{8}{10}\selectfont{\textrm{{Topic ID}}} & \fontsize{8}{10}\selectfont{\textrm{{Top 10 Salient words}}} & \fontsize{8}{10}\selectfont{\textrm{{Token Contribution}}} & \fontsize{8}{10}\selectfont{\textrm{{Inferred Topic}}}\\
		\hline
		\fontsize{8}{10}\selectfont{\textrm{{1}}} & \fontsize{8}{10}\selectfont{\textrm{{M-pox, vaccine, travel, supply, vaccination}}} & \fontsize{8}{10}\selectfont{\textrm{{33.5$\%$}}} & \fontsize{8}{10}\selectfont{\textrm{{Vaccine Safety and}}}\\
		\fontsize{8}{10}\selectfont{\textrm{{}}} & \fontsize{8}{10}\selectfont{\textrm{{side, effect, response, safe, prevent}}} & \fontsize{8}{10}\selectfont{\textrm{{}}} & \fontsize{8}{10}\selectfont{\textrm{{Availability Concerns}}}\\
		\fontsize{8}{10}\selectfont{\textrm{{2}}} & \fontsize{8}{10}\selectfont{\textrm{{cases, M-pox, deadly, sores, blisters}}} & \fontsize{8}{10}\selectfont{\textrm{{22.9$\%$}}} & \fontsize{8}{10}\selectfont{\textrm{{Fear of Death from}}}\\
		\fontsize{8}{10}\selectfont{\textrm{{}}} & \fontsize{8}{10}\selectfont{\textrm{{death, recovery, harmless, population, rate}}} & \fontsize{8}{10}\selectfont{\textrm{{}}} & \fontsize{8}{10}\selectfont{\textrm{{ M-pox Infection}}}\\
		\fontsize{8}{10}\selectfont{\textrm{{3}}} & \fontsize{8}{10}\selectfont{\textrm{{Pox, monkey, transmit, airborne, mask}}} & \fontsize{8}{10}\selectfont{\textrm{{20.3$\%$}}} & \fontsize{8}{10}\selectfont{\textrm{{Possible Modes of}}}\\
		\fontsize{8}{10}\selectfont{\textrm{{}}} & \fontsize{8}{10}\selectfont{\textrm{{water, hand, touch, animal, catch}}} & \fontsize{8}{10}\selectfont{\textrm{{}}} & \fontsize{8}{10}\selectfont{\textrm{{M-pox Transmission}}}\\
		\fontsize{8}{10}\selectfont{\textrm{{4}}} & \fontsize{8}{10}\selectfont{\textrm{{Covid, new,  M-pox, mutation, lies,}}} & \fontsize{8}{10}\selectfont{\textrm{{11.7$\%$}}} & \fontsize{8}{10}\selectfont{\textrm{{Conspiracy Theories about}}}\\
		\fontsize{8}{10}\selectfont{\textrm{{}}} & \fontsize{8}{10}\selectfont{\textrm{{media, fake, mutation, corruption, greed}}} & \fontsize{8}{10}\selectfont{\textrm{{}}} & \fontsize{8}{10}\selectfont{\textrm{{the M-pox outbreak}}}\\
		\fontsize{8}{10}\selectfont{\textrm{{5}}} & \fontsize{8}{10}\selectfont{\textrm{{Pox, covid, new, gay, sex, racist}}} & \fontsize{8}{10}\selectfont{\textrm{{11.6$\%$}}} & \fontsize{8}{10}\selectfont{\textrm{{M-pox related Stigmatisation}}}\\
		\fontsize{8}{10}\selectfont{\textrm{{}}} & \fontsize{8}{10}\selectfont{\textrm{{monkey, African, black, border}}} & \fontsize{8}{10}\selectfont{\textrm{{}}} & \fontsize{8}{10}\selectfont{\textrm{{and Discrimination}}}\\
			\hline
		\fontsize{8}{10}\selectfont{\textrm{{Topic ID}}} & \fontsize{8}{10}\selectfont{\textrm{{Top 10 Salient words}}} & \fontsize{8}{10}\selectfont{\textrm{{Token Contribution}}} & \fontsize{8}{10}\selectfont{\textrm{{Inferred Topic}}}\\
		\hline
		\fontsize{8}{10}\selectfont{\textrm{{1}}} & \fontsize{8}{10}\selectfont{\textrm{{M-pox, virus, covid, spread, gay}}} & \fontsize{8}{10}\selectfont{\textrm{{27.6$\%$}}} & \fontsize{8}{10}\selectfont{\textrm{{M-pox as a Sexually}}}\\
		\fontsize{8}{10}\selectfont{\textrm{{}}} & \fontsize{8}{10}\selectfont{\textrm{{sex, condom, risk, apply, concern}}} & \fontsize{8}{10}\selectfont{\textrm{{}}} & \fontsize{8}{10}\selectfont{\textrm{{Transmitted Disease (STD)}}}\\
		\fontsize{8}{10}\selectfont{\textrm{{2}}} & \fontsize{8}{10}\selectfont{\textrm{{M-pox, vaccine, side, effects, safe}}} & \fontsize{8}{10}\selectfont{\textrm{{23.4$\%$}}} & \fontsize{8}{10}\selectfont{\textrm{{Vaccine Safety and}}}\\
		\fontsize{8}{10}\selectfont{\textrm{{}}} & \fontsize{8}{10}\selectfont{\textrm{{available, supply, smallpox, outbreak, death}}} & \fontsize{8}{10}\selectfont{\textrm{{}}} & \fontsize{8}{10}\selectfont{\textrm{{Availability Concerns}}}\\
		\fontsize{8}{10}\selectfont{\textrm{{3}}} & \fontsize{8}{10}\selectfont{\textrm{{M-pox, covid, hoax, mask, variant}}} & \fontsize{8}{10}\selectfont{\textrm{{19.5$\%$}}} & \fontsize{8}{10}\selectfont{\textrm{{Conspiracy Theories about}}}\\
		\fontsize{8}{10}\selectfont{\textrm{{}}} & \fontsize{8}{10}\selectfont{\textrm{{outbreak, variant, global, cause, want}}} & \fontsize{8}{10}\selectfont{\textrm{{}}} & \fontsize{8}{10}\selectfont{\textrm{{the M-pox Outbreak}}}\\
		\fontsize{8}{10}\selectfont{\textrm{{4}}} & \fontsize{8}{10}\selectfont{\textrm{{Pox, monkey, get, real, scary}}} & \fontsize{8}{10}\selectfont{\textrm{{17.2$\%$}}} & \fontsize{8}{10}\selectfont{\textrm{{Fear of M-pox related}}}\\
		\fontsize{8}{10}\selectfont{\textrm{{}}} & \fontsize{8}{10}\selectfont{\textrm{{sores, ugly, picture, cream, scratch}}} & \fontsize{8}{10}\selectfont{\textrm{{}}} & \fontsize{8}{10}\selectfont{\textrm{{skin lesions and scarring}}}\\
		\fontsize{8}{10}\selectfont{\textrm{{5}}} & \fontsize{8}{10}\selectfont{\textrm{{New, infection, pandemic, shutdown}}} & \fontsize{8}{10}\selectfont{\textrm{{12.3$\%$}}} & \fontsize{8}{10}\selectfont{\textrm{{Potential Emergence of a}}}\\
		\fontsize{8}{10}\selectfont{\textrm{{}}} & \fontsize{8}{10}\selectfont{\textrm{{M-pox, kill, death, again, global, public}}} & \fontsize{8}{10}\selectfont{\textrm{{}}} & \fontsize{8}{10}\selectfont{\textrm{{deadly M-pox pandemic}}}\\
		\hline
 \end{tabular}}}
        \caption{\fontsize{8}{10}\selectfont{\textrm{{\underline{A Comparison of the LDA Results of the Mislabelled M-pox tweets by the RoBERTa}}}}}\hspace{1.25cm}{\fontsize{8}{10}\selectfont{\textrm{{\underline{model before (top-section) and after (bottom-section) being fine-tuned.}}}}}

\end{table}
\fontsize{9}{12}\selectfont{\textrm{{\textrm{{Here we contrast the results of the LDAs of the RoBERTa model pre- and post- training on the mpox dataset, in which the number of 5 topics was set to five. The size of the data-sets were dependant on the predictive power or accuracy of the model on the full set of M-pox tweets (20604), in which pre-training the accuracy was 47$\%$ on the full set of tweets (20604 tweets) and the accuracy post-training was 69$\%$ on the testing sample which contributed 1/5 or one-fifth of the entire sample (2843 tweets).}}}}

\fontsize{9}{12}\selectfont{\textrm{{\textrm{The top 10 most salient words, the associated inferred topic and the document contribution provided per topic. From the results, one can see that the set of topics in both LDAs involve discussions or themes pertaining to vaccination, transmission, fear and panic, recovery, treatment as well as accompanying conspiracy theories.}}}}

\fontsize{9}{12}\selectfont{\textrm{{\textrm{With respect to the LDA for the RoBERTa model prior to training, the topic with the highest contribution was ``Vaccine Safety and Availability Concerns'' at $\approx$ 34$\%$, which also happened to be common to both LDA reports and also the second highest topic for the RoBERTa model post-training at $\approx$ 24$\%$. The only other topic that was identical in both LDA analyses was given by: ``Conspiracy Theories about the M-pox Outbreak'', which contributed $\approx$ 12$\%$ pre-training, but $\approx$ 20$\%$ post-training.}}}}
\fontsize{9}{12}\selectfont{\textrm{{\textrm{The topic with the lowest contribution pre-training was taken by ``M-pox related Discrimination and Stigmatisation" at $\approx$ 12$\%$, which correlates  to the topic of lowest contribution for the case of the RoBERTa model post-training, since this discrimination and stigmatisation is closely related or proportional to the degree of fear of M-pox being the next pandemic, given by the title, ``Potential Emergence of a deadly M-pox pandemic'', also at $\approx$ 12$\%$.}}}}

\fontsize{9}{12}\selectfont{\textrm{{\textrm{It is apparent that although 3 out of the 5 topics in the one set of LDA results are not identical to the other, that they are in essence the same entity or grouping i.e. topics within a topic.}}}} 

\fontsize{9}{12}\selectfont{\textrm{{\textrm{ For example,``Possible Modes of M-pox Transmission'' and ``M-pox as an STD.'', are both sub-topics of the general topic of M-pox transmission and infectivity; ``Fear of M-pox related skin lesions and scarring" and "Fear of death by M-pox", are both sub-topics of the general topic of M-pox infection and recovery; ``Potential Emergence of an M-pox pandemic'' and ``M-pox related Stigmatization and Discrimination'', are both sub-topics of the general topic of public mass panic and hysteria.}}}} 

\fontsize{9}{12}\selectfont{\textrm{{\textrm{It appears that, after training the model became better (decrease in topic contribution) at correctly labelling topics referring to problematic M-pox symptoms by $\approx$ 6$\%$, better at predicting correct labels pertaining to vaccination by $\approx$ 10$\%$, but worse at predicting the labels for tweets pertaining to conspiracy theories by $\approx$ 8$\%$ - assuming that the smaller testing sample was representative of the entire sample in terms of thematic content.}}}}

\fontsize{9}{12}\selectfont{\textrm{{\textrm{Nevertheless, this hints towards an overlap in terms of themes present in the respective topics - especially when it comes to vaccination }}}}}
\section*{\fontsize{10}{13}\selectfont{\textrm{Acknowledgments}}}
\fontsize{10}{13}\selectfont{\textrm{We give a big thank you to Canada’s International Development Research Centre (IDRC) and the Swedish Inter- national Development Cooperation Agency (SIDA) (Grant No. 109559-001) for funding this research.}}

\end{document}